\documentclass[sn-mathphys,Numbered]{sn-jnl}

\usepackage{graphicx}%
\usepackage{multirow}%
\usepackage{amsmath,amssymb,amsfonts}%
\usepackage{amsthm}%
\usepackage{mathrsfs}%
\usepackage[title]{appendix}%
\usepackage{xcolor}%
\usepackage{textcomp}%
\usepackage{manyfoot}%
\usepackage{booktabs}%
\usepackage{algorithm}%
\usepackage{algorithmicx}%
\usepackage{algpseudocode}%
\usepackage{listings}%

\usepackage{float} 
\usepackage{subcaption}
\usepackage{bm}
\usepackage{tabularx}
\theoremstyle{thmstyleone}%
\theoremstyle{thmstyletwo}%

\theoremstyle{thmstylethree}%

\newenvironment{sequation}{\begin{equation}\small}{\end{equation}}

\begin{document}

\title[Article Title]{De-confounding Representation Learning for Counterfactual Inference on Continuous Treatment via Generative Adversarial Network}


\author[1]{\fnm{Yonghe} \sur{Zhao}} 

\author[1]{\fnm{Qiang} \sur{Huang}} 

\author[1]{\fnm{Haolong} \sur{Zeng}} 

\author[2]{\fnm{Yun} \sur{Peng}} 

\author*[1]{\fnm{Huiyan} \sur{Sun}} 

\affil*[1]{\orgdiv{School of Artificial Intelligence}, \orgname{Jilin University}, \orgaddress{\street{Qianjin Street}, \city{Changchun}, \postcode{130012}, \state{Jilin}, \country{China}}}

\affil[2]{\orgdiv{Department of Data Analysis}, \orgname{Baidu}, \orgaddress{\street{Shangdi Street}, \city{Beijing}, \postcode{100085}, \country{China}}}


\abstract{Counterfactual inference for continuous rather than binary treatment variables is more common in real-world causal inference tasks. While there are already some sample reweighting methods based on Marginal Structural Model for eliminating the confounding bias, they generally focus on removing the treatment’s linear dependence on confounders and rely on the accuracy of the assumed parametric models, which are usually unverifiable. In this paper, we propose a de-confounding representation learning (DRL) framework for counterfactual outcome estimation of continuous treatment by generating the representations of covariates disentangled with the treatment variables. The DRL is a non-parametric model that eliminates both linear and nonlinear dependence between treatment and covariates. Specifically, we train the correlations between the de-confounded representations and the treatment variables against the correlations between the covariate representations and the treatment variables to eliminate confounding bias. Further, a counterfactual inference network is embedded into the framework to make the learned representations serve both de-confounding and trusted inference. Extensive experiments on synthetic datasets show that the DRL model performs superiorly in learning de-confounding representations and outperforms state-of-the-art counterfactual inference models for continuous treatment variables. In addition, we apply the DRL model to a real-world medical dataset MIMIC \uppercase\expandafter{\romannumeral3} and demonstrate a detailed causal relationship between red cell width distribution and mortality.}




\maketitle

\section{Introduction}\label{sec1}
\begin{figure*}[ht]
\centering
\includegraphics[width=0.8\textwidth]{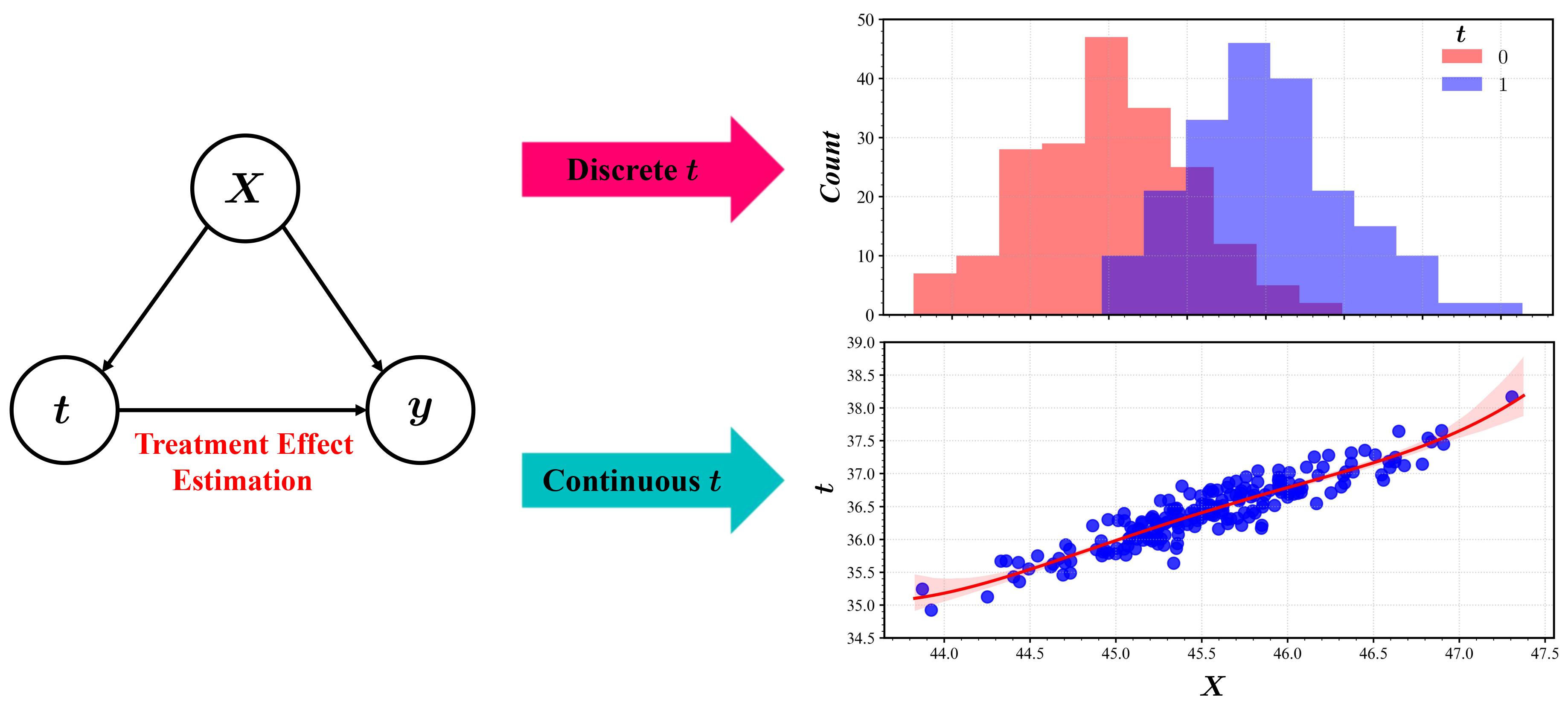}
\caption{The issues engendered by confounding covariates are twofold: (i) inconsistent distribution of $X$ amidst discrete $t$ values; (ii) a distributive interdependence between $X$ and continuous $t$ values.}
\label{confounders}
\end{figure*}

Rational counterfactual inference from the observational data are essential for decision making\cite{Rubin1974Estimating}. For example, the choice of medical options for a patient\cite{2019MachineMed}, the evaluation of the actual effectiveness of an economic measure\cite{learning}, or the availability of a new vaccine\cite{destefano2007vaccines}, etc. Where the primary focus of this paper lies in the subdivision field for counterfactual inference of continuous treatment. In practice, continuous treatment variables, including but not limited to drug dosage for patients\cite{imbens2000role}, strength of government economic subsidies\cite{kluve2012evaluating}, and amount of political or commercial advertising\cite{2018Covariate}, are frequently encountered. A consensus is that collecting data from prospectively designed experiment, called randomized controlled trial (RCT), is the gold standard for counterfactual inference from observational data\cite{JudeaCausality}. However RCTs are time-consuming and expensive, even involving ethical issues in some scenarios\cite{2010Reputation,2016Assessing,2011Unexpected}. Different from the randomness of treatments in RCTs, the main challenge of causal inference in observational studies is the unknown mechanism of treatment assignment. That is, there exist covariates that influence both treatment and outcome variables, commonly referred to as confounders\cite{imbens2015causal}. Specifically, as shown in Fig. \ref{confounders}, the covariates $X$ affect the selection of treatment $t$ thus leading to: (i) inconsistent distribution of $X$ amidst discrete $t$ values; (ii) a distributive interdependence between $X$ and continuous $t$ values. Further, these phenomena result in unsatisfactory accuracy of counterfactual inference, which are similar to the domain adaptation problem or collinearity of covariates\cite{yao2020survey}.

Various models for de-confounding of observed confounders are proposed under the unconfoundedness assumption, which refers to the absence of unobserved confounders\cite{imbens2015causal}, including reweighting\cite{rosenbaum1983central,2019Robust,lee2011weight,Austin2011An,imai2014covariate}, matching\cite{L2021Combining,2017Informative,JMLR21}, causal trees\cite{2010BART}, confounding balanced representation learning\cite{Weighted,learning,perfect,2019AdversarialDu}, etc. However, most of these models target discrete or even binary treatment and are not sufficiently scalable for continuous treatment variables. The primary reason for this disconnection is that de-confounding models for discrete treatments focus on balancing confounders across different treatment groups. However, for continuous treatments, such groups are dense and infinite, which makes achieving balance a much more challenging task. Indeed, de-confounding algorithms for continuous treatment have attracted a widespread attention, mainly including inverse conditional probability-of-treatment weighting methods based on marginal structural equations\cite{2000Marginal,2015A,2018Covariate,2021continuous}, double robust weighting adjustment methods\cite{drref} and deep discretization methods for continuous treatment\cite{2019LearningC,SCIGAN,zou2020learning}, etc. Where, the most reweighting methods on continuous treatment focus on linear allocation mechanisms, correlation of treatment and covariates, and causal relationships. Additionally, these models rely on the accuracy of the assumed parametric models. There are two key issues in the above setting: first, linearity is merely a specific and simplistic form of nonlinearity; and second, both the allocation mechanism and the causality remain unclear, making the accuracy of the parametric models unverifiable. Meanwhile, Deep discretization methods for multi-treatment dose-response curves are still fundamentally de-confounding algorithms at the discrete treatment level.

In this paper, we propose a counterfactual inference model DRL centred on acquiring de-confounding representations of covariates. Specifically, we first use random sampling to generate virtual de-confounding representations in a pre-defined representation space and treat their correlations with the treatment variables as samples of the real distribution. Then, we perform adversarial training on the correlations between to-be-learned representations of covariates and treatment variables against the real distribution. Further, a counterfactual inference network is embedded into the framework to perform trusted counterfactual inference. The main contributions of this paper are summarized as follows.
\begin{itemize}
\item We propose a novel representation learning framework DRL to eliminate the confounding bias for counterfactual inference of continuous treatment, which is a non-parametric model, adaptively balances the confounders by adversarial networks to avoid the bias of artificially selected parameters.
\item The DRL framework presents a comprehensive nonlinear perspective, distinct from prevailing counterfactual inference models for continuous treatment variables, across three crucial domains: the mechanism of treatment assignment, the correlation between treatment and covariates, and the causal relationship linking treatment and outcome variables.
\item Extensive experiments on the synthetic datasets illustrate that the DRL is significantly superior to the state-of-the-art models in both de-confounding and counterfactual inference for continuous treatments. We apply the proposed DRL to a real-world dataset MIMIC \uppercase\expandafter{\romannumeral3} and identify a detailed causal relationship between red cell width distribution and mortality, which contributes to precision medicine.
\end{itemize}
\section{Related Works}
Since continuous treatment variables differ from the natural hierarchical properties of discrete variables, various methods for discrete treatment are unable to directly deal with the continuous treatment. The classical method for counterfactual inference of continuous treatment is through Marginal Structural Models\cite{2000Marginal,imbens2004nonparametric}, which are based on inverse conditional probability-of-treatment weights (ICPW). The ICPW mothod assumes a linear relationship between the outcome and the treatment variables and establishes the marginal structural equation, as shown in Eq.~(\ref{equ1}). To eliminate the effects of confounders, the ICPW method advocates reweighting the original samples with inverse conditional probability-of-treatment weights, as shown in Eq.~(\ref{equ2}), and then solving for the parameters $\alpha_{0}$ and $\alpha_{1}$ in Eq. (\ref{equ1}).
\begin{equation}
\mathbb{E}[y(t)] = \alpha_{0} + \alpha_{1}t
\label{equ1}
\end{equation}
\begin{equation}
\label{equ2}
w_{i} = \frac{f_{t}(t_{i})}{f_{t|X}(t_{i}|X_{i})},\ i = 1,2,\cdots,N
\end{equation}
where, $f_{t}$ represents the probability density function of the treatment variables and $f_{t|X}$ represents the conditional probability density of the treatment variables with respect to the covariates. The ICPW method assumes a linear relationship among covariates, treatment and outcome variables. However, the accuracy of ICPW model depends on the correct definition of the parametric functions of conditional probability densities and marginal structural equation, which are typically unverifiable.

Based on the idea of ICPW, various counterfactual inference algorithms based on weight-adjusted for continuous treatment variables have emerged. The GBM model\cite{2015A} uses a boosting algorithm to estimate the nonlinear conditional probability density of the treatment variables, thus improving on the assumption of linear correlation between covariates and treatment variables in ICPW. Meanwhile, the GBM model proposes a method to determine the optimal number of decision trees in the boosting algorithm based on the target of reducing the linear correlation between treatment variables and covariates. Nevertheless, the boosting algorithm does not guarantee the accuracy of the conditional probability density estimation due to the unidentified assignment mechanism. To address this issue, the CBGPS model\cite{2018Covariate} advocates learning the weights with the criterion of de-confounding as the target, unlike ICPW and GBM, which divide the construction of weights and de-confounding into two steps. CBGPS allows the weights to directly serve the objective of de-confounding, thereby eliminating concerns regarding the correct definition of the conditional probability density form. Similarly, Kallus and Santacatterina proposed a non-parametric approach KOOW based on convex optimization\cite{kallus2019kernel}, which aims to minimize the covariance of the worst-case penalty function between the continuous treatment and confounding factors by optimizing the sample weight. Unfortunately, the CBGPS and KOOW only consider reducing the linear correlation between the treatment variables and the covariates. Further, the GAD method\cite{2021continuous} proposes to learn a sample weight on the observational data through adversarial networks, such that the distribution of weighted observational data would be similar even identical with the “calibration” data obtained by random permutation. GAD acquires covariates dissociated with the treatment variables by randomly disrupting the covariates, thereby eliminating both linear and nonlinear correlations between the treatment variables and the covariates to some extent. However, learning sample weights through adversarial networks over the entire sample space is time-consuming and expensive. In addition, the above methods do not explore the nonlinear causal relationship between covariates and outcome variables. Furthermore, the poor scalability of reweighting methods for new samples has not received sufficient attention.

In practice, some researchers compromise by discretizing continuous treatment based on deep learning methods. The DRNets model\cite{2019LearningC} discretizes the continuous dose into different intervals to develop a multi-head neural network structure. Meanwhile, SCIGAN\cite{SCIGAN} constructs an end-to-end multi-task adversarial framework to infer counterfactual outcomes by sampling over continuous treatment. Although the above models propose corresponding solution for determining the number of intervals or samples, an accurate counterfactual inference model for continuous treatment is still a further demand. In this paper, we propose DRL model based on representation learning and adversarial training to adaptively learn representations of covariates independent of the treatment variables, while fully considering the nonlinear relationships among covariates, treatment and outcome variables.

\section{Method}
\subsection{Symbol Description} 
In this paper, $X$ represents the original covariates, $X^{R}$ denotes the randomly generated virtual representations in the representation space that are idealized disentangled with the treatment $t \in \mathcal{T}$, and $X^{G}$ denotes the de-confounding representations of $X$ that are intended to be learned. For the outcome variable $y \in \mathcal{Y}$, only the factual outcome $y_{i}^{f}(t_{i})$ corresponding to $t_{i}$ is observable in practice. While the counterfactual outcomes $y_{i}^{cf}(C_{\mathcal{T}}t^{i})$ are not accessible, where $C_{\mathcal{T}}t_{i}$ represent the complement of $t_{i}$ with respect to $\mathcal{T}$. The proposed method aims to perform counterfactual inference $y_{i}^{cf}(C_{\mathcal{T}}t^{i})$ using the learned de-confounding representations $X^{G}$ and continuous treatment $t$.
\subsection{Assumption} 
According to the potential outcome framework, the DRL model necessitates the fulfillment of three assumptions: stable unit treatment value (SUTV), unconfoundedness, and positivity assumption\cite{rosenbaum1983central,imbens2015causal}.

The SUTV assumption includes: firstly, the potential outcome of each individual is not affected by the treatment of any other individual, in other words, individuals are independent; secondly, there is no measurement error in the factual observational outcome. 

The unconfoundedness assumption represents that the treatment variable is independent of the outcome variable given the covariates $X$, i.e., $\mathcal{T}\perp\mathcal{Y}|X$. With this unconfoundedness assumption, for the samples with the same covariates $X$, their treatment assignment can be viewed as random.

The positivity assumption, commonly referred to as the overlap assumption, posits that each value of $X$ can be assigned to any treatment with a non-zero probability, specifically $p(t|X=x) >0, \forall\ t \in \mathcal{T}, x \in X$. The purpose of counterfactual inference is to assess differences across treatments, and the model is meaningless if some treatments can not be observed or are not meaningful. 
\begin{figure*}[h]
  \centering
  \includegraphics[width=\linewidth]{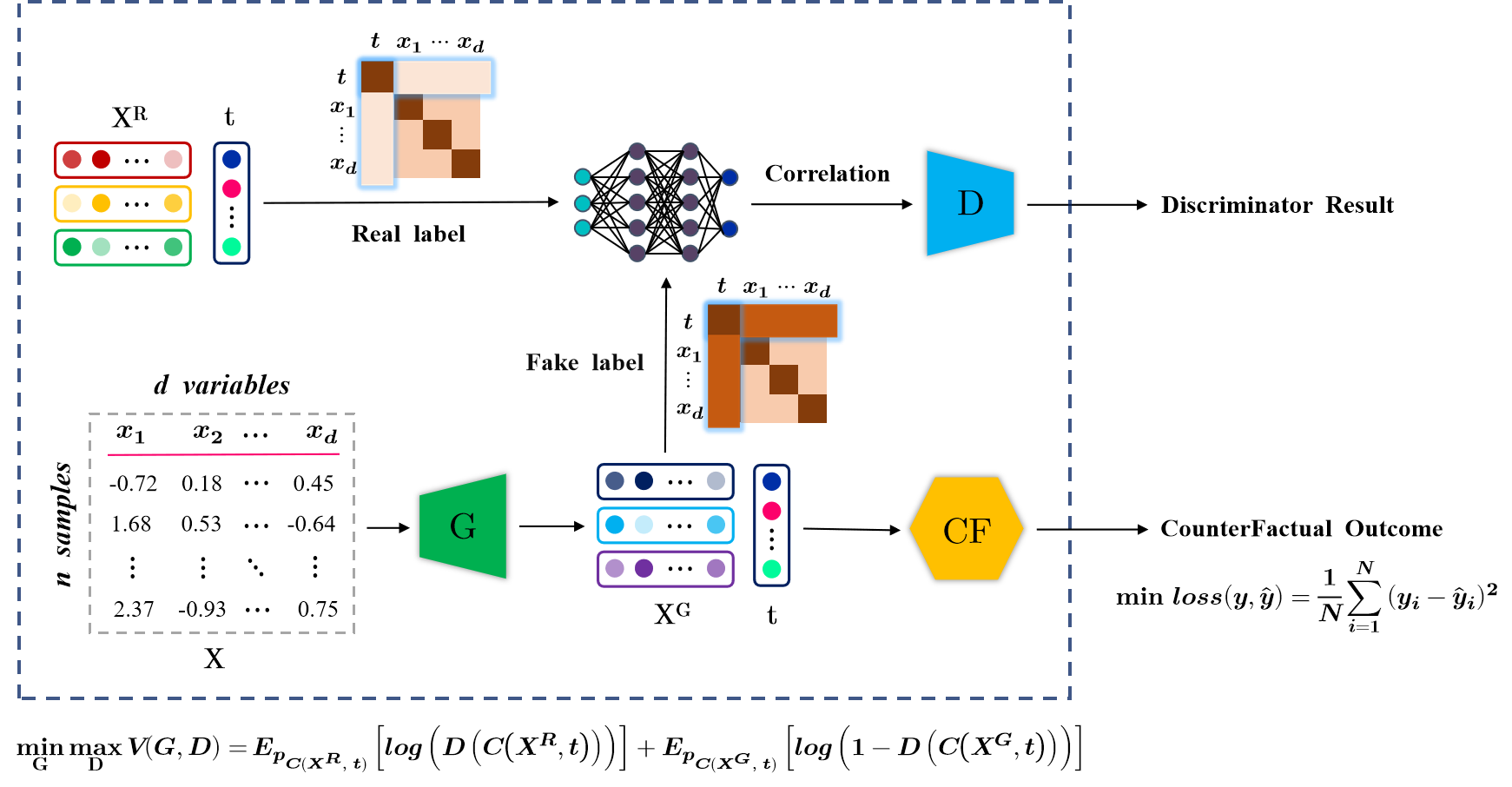}
  \caption{The overview framework of the de-confounding representation learning that contains four sub-modules: Generator, Discriminator, Correlation network and CounterFactual module.}
  \label{Architecture}
\end{figure*}
\subsection{De-confounding Representation Learning}
\subsubsection{Model Structure} 
The DRL model is based on the principle of adversarial training, which aims to acquire representations of covariates that are disentangled from the treatment variables in order to facilitate counterfactual inference. The model posits two complementary objectives that mutually reinforce each other, namely de-confounding representation learning and counterfactual inference.

The overall framework of the DRL model is designed to address the aforementioned objectives, as illustrated in Fig.~\ref{Architecture}. Specifically, the DRL employs an adversarial network, comprised of the Generator, Correlation network and Discriminator in Fig. \ref{Architecture}, to achieve de-confounding representation learning. Initially, virtual representations $X^{R}$ are randomly generated in the representation space of a specific dimension, while the Generator produces the representations $X^{G}$ of the original covariates to be learned. Next, the Correlation network takes covariate representations ($X^{R}$ or $X^{G}$) and treatment variables $t$ as input to the interaction and extracts their correlation. The correlation between the $X^{R}$ and the treatment variables $t$ is utilized as the real data, meanwhile the correlation between the to-be-learned $X^{G}$ and $t$ is leveraged as the fake data for adversarial training of the Discriminator. If the Discriminator is unable to differentiate between the virtual and to-be-learned representations, it suggests that the learned $X^{G}$ are likely to be disentangled from the treatment variables. Furthermore, to ensure that the learned representations are effective in inferring counterfactual outcomes, $X^{G}$ and the treatment variables $t$ are fed into the CounterFactual module in Fig. \ref{Architecture} for training to minimize inference errors.

\subsubsection{Model Principle}
The initial motivation for Generative Adversarial Network (GAN) is to learn the real distribution underlying the observational data\cite{2014Generative}. However, the DRL model places more emphasis on the correlation between the representations ($X^{R}$ or $X^{G}$) and the treatment variables $t$. In DRL, the Correlation network extracts correlations between the representations ($X^{R}$ or $X^{G}$) and the $t$. Therefore, the objective of adversarial training is to analyze the correlation between $X^{R}$ and $X^{G}$ with the $t$, rather than analyzing $X^{R}$ and $X^{G}$ individually.

Most de-confounding models for continuous treatment artificially define one or more objectives to measure the correlation between treatment variables and covariates to guide the training process. For instance, these objectives may include the Pearson correlation coefficient being equal to zero, or the expectation of the product of treatment variables and covariates being equal to the product of their respective expectations. However, these objectives have certain limitations, as they only measure the linear correlation between variables. In this paper, since the virtual representations as the real data for adversarial training are randomly generated in the representation space, $X^{R}$ is naturally independent of the treatment variable at both the linear and nonlinear levels. Additionally, the Correlation network adapts to learn the criterion for measuring relevance and the Discriminator makes a judgment as to whether the input data meets the standard. Consequently, during the adversarial process, the Discriminator gradually specifies the correlations between continuous treatment and representations, and the Generator ultimately generates representations that are disentangled from the continuous treatment. 

Moreover, we incorporate the CounterFactual module into the process of optimizing the Generator, which enables the generated representations to better serve the objective of the final counterfactual inference. Naturally, generating representations that are disentangled from the continuous treatment improves the accuracy of counterfactual prediction. In turn, the objective of CounterFactual module also guides the selection of more valid representations for counterfactual outcomes.
\subsubsection{Optimization Objectives} 
To define the objective functions of the DRL model, we first refer to the objective functions of the GANs model as shown in Eq. (\ref{equ6})\cite{2014Generative}. Where $G$ and $D$ represent the generator and discriminator, respectively; $G(z)$ represents the generated fake data, and $D(x)$ represents the discriminative result of data $x$. Goodfellow et al. demonstrated that the optimal solution of the GANs model is achieved when $p(x) = p(G(z))$, and when the output value of the Discriminator is 0.5, indicating that it is incapable of distinguishing between real and fake data.
\begin{equation}
\label{equ6}
\mathop{min}\limits_{G}\mathop{max}\limits_{D}V(G,D) = \mathbb{E}_{p_{x}}[log(D(x)] + \mathbb{E}_{p_{z}}[log(1-D(G(z)))]
\end{equation}

Similarly, the objectives of the DRL model is to learn the representation of covariates that are disentangled from the treatment variables. Where the correlation $C(X^{R},t)$ between the $X^{R}$ and the $t$ is treated as the real data and the correlation $C(X^{G},t)$ between the to-be-learned $X^{G}$ and $t$ as the generated fake data, the corresponding objective function is shown in Eq. (\ref{equ7}).
\begin{sequation}
\label{equ7}
\mathop{min}\limits_{G}\mathop{max}\limits_{D}V(G,D) = \mathbb{E}_{p_{C(X^{R},t)}}[log(D(C(X^{R},t)))] + \mathbb{E}_{p_{C(X^{G},t)}}[log(1-D(C(X^{G},t)))]
\end{sequation}

In addition to the adversarial training, the counterfactual inference error shown in Eq. (\ref{equ8}) is also a part of the overall objectives. When the outcome variable is in discrete form, the counterfactual inference error is shown in Eq. (\ref{equ12}), where $y_{i}^{c}$ represents the $i$-th outcome that belongs to category $c$.
\begin{equation}
\label{equ8}
\mathop{min}\mathop{loss}(y,\hat{y}) = \frac{1}{N}\sum_{i=1}^{N}(y_{i}-\hat{y}_{i})^{2}
\end{equation}
\begin{equation}
\label{equ12}
\mathop{min}\mathop{loss}(y,\hat{y}) = -\frac{1}{N}\sum_{i=1}^{N}\sum_{c=1}^{C}y_{i}^{c}log(\hat{y}_{i}^{c})
\end{equation}
\subsubsection{Optimising Processes} 
Based on the observational dataset $\mathcal{D} = \{X^{i},t^{i},y^{i}\}_{i = 1}^{N}$, virtual representations $X^{R}$ are randomly generated in the specified dimension of the representation space. The proposed model is then trained in three incremental steps. Firstly, with the coefficients of the Generator and CounterFactual modules fixed, the Correlation network and Discriminator and  are trained using the objective function shown in Eq. (\ref{equ9}), which is the maximization objective in Eq. (\ref{equ7}). During this step, the Correlation network learns a data-driven criterion for measuring the correlation between the representation of covariates and treatment variables, and the Discriminator discriminates whether the input data meets that criterion.
\begin{equation}
\label{equ9}
\mathop{max} l^{d} = \mathbb{E}_{p_{C(X^{R},t)}}[log(D(C(X^{R},t)))] + \mathbb{E}_{p_{C(X^{G},t)}}[log(1-D(C(X^{G},t)))]
\end{equation}

Secondly, keeping the coefficients of the Correlation network, Discriminator and CounterFactual modules fixed, the coefficients of the Generator are optimized based on the Eq. (\ref{equ10}), namely minimization objective in Eq. (\ref{equ7}) and the counterfactual inference error in Eq. (\ref{equ8}). As shown in Eq. (\ref{equ10}), the step of optimizing the Generator takes into account not only the correlation between the representation and the treatment variables, but also the accuracy of the counterfactual inference. It is noteworthy that the objective function $l^{g}$ includes weighting factors $w^{c}$, which control the degree of influence of the counterfactual error on the representation generation.
\begin{equation}
\label{equ10}
\mathop{min} l^{g} = \mathbb{E}_{p_{C(X^{G},t)}}[log(1-D(C(X^{G},t)))] + w^{c}\mathop{loss}(y,\hat{y})
\end{equation}

Finally, the CounterFactual module is optimized based on the objective function $l^{c}$, as shown in Eq. (\ref{equ11}). During this optimization step, the coefficients of the Generator, Correlation network and the Discriminator are kept constant to prevent negative influence during training between the different steps.
\begin{equation}
\label{equ11}
\mathop{min} l^{c} = \mathop{loss}(y,\hat{y})
\end{equation}

The above three steps are cyclic and mutually reinforcing, working together to achieve the two predefined objectives: de-confounding representation learning and counterfactual inference.

\section{Experiments}
\subsection{Datasets}
Since counterfactual outcomes cannot be collected beyond the observational data in practice, some synthetic or semi-synthetic datasets are generally applied to evaluate the performance of the counterfactual inference model. In order to evaluate the DRL model comprehensively, we synthesize experimental data for four scenarios: the combinations of linear and nonlinear treatment assignment and outcomes generation.

Covariates, which are considered exogenous variables in the potential outcome framework, are firstly randomly generated by multivariate Gaussian distribution, as shown in Eq.~(\ref{equ13}), where $d = 10$ in the experiment.
\begin{equation}
\label{equ13}
X \sim N(0^{1\times d},\frac{1}{2}(\Sigma + \Sigma^{T})), \Sigma \sim U((-1,1)^{d\times d})
\end{equation}
Then, based on the generated covariates, treatment and outcome variables are generated from Eq.~(\ref{equ14}). Where, random error $\epsilon_{t} \sim N(0,0.3)$, $\epsilon_{y} \sim N(0,0.5)$.
\begin{equation}
\label{equ14}
t = f_{t}(X) + \epsilon_{t};\ y = f_{y}(X) + f_{y}(t) +\epsilon_{y}
\end{equation}
Specifically, $f_{t}$ and $f_{y}$ have different forms of generating mechanisms regarding the linear or nonlinear $t$ and $y$. The linear and nonlinear $t$ are generated by $f_{t}^{line}$ and $f_{t}^{nonL}$ shown in Eq.~(\ref{equ16}). Where, the parameters $w^{xt} \sim U(1,5)$.
\begin{equation}
\label{equ16}
f_{t}^{line}(X) = \sum_{i=1}^{d}w_{i}^{Xt}X_{i};\ f_{t}^{nonL}(X) = sigmoid(\sum_{i=1}^{d}w_{i}^{Xt}X_{i})
\end{equation}
Similarly, for linear $y$, $f_{y}^{line}$ is shown in Eq.~(\ref{equ18}) and nonlinear $y$ is defined by Eq.~(\ref{equ19}). Where, parameters $w^{xy} \sim U(1,5)$, $w^{ty} = 5$.
\begin{equation}
\label{equ18}
f_{y}^{line}(X) = \sum_{i=1}^{d}w_{i}^{Xy}X_{i};\ f_{y}^{line}(t) = w^{ty}t
\end{equation}
\begin{equation}
\label{equ19}
f_{y}^{nonL}(X) = sigmoid(\sum_{i=1}^{d}w_{i}^{Xy}X_{i});\ f_{y}^{nonL}(t) = sigmoid(w^{ty}t)
\end{equation}

The contrast experiments are conducted in the four scenarios which are specifically distinguished based on the respective two generation methods of $f_{t}$ and $f_{y}$, including Scenario A: $f_{t}^{line}$ and $f_{y}^{line}$; Scenario B: $f_{t}^{line}$ and $f_{y}^{nonL}$; Scenario C: $f_{t}^{nonL}$ and $f_{y}^{line}$; Scenario D: $f_{t}^{nonL}$ and $f_{y}^{nonL}$. 
\subsection{Evaluation Metrics} 
The central target of the DRL model focuses on de-confounding and counterfactual inference for continuous treatment variables. Therefore, the evaluation metrics of the experiments are designed to center around measuring correlation and counterfactual inference accuracy. 

Regarding correlation, in contrast to most causal models for continuous treatment that concentrate solely on the linear correlation between the treatment variables and covariates, we additionally emphasize the index of nonlinear correlation. Typically, the average Pearson Correlation Coefficient (PCC) is utilized to gauge the linear correlation between the treatment variables and the covariates. In addition, since the covariates are a set of variables, we report the Multiple Correlation Coefficient (MCC) shown in Eq.~(\ref{equ20}), a metric that assesses the extent of correlation between a variable and a set of variables\cite{1991MULTIPLE}, to illustrate the de-confounding performance on both linear and nonlinear level.
\begin{equation}
\label{equ20}
{\rm MCC}_{Xt} = \frac{\sum(t-\bar{t})(\hat{t}-\bar{t})}{\sqrt{\sum(t-\bar{t})^{2}\sum(\hat{t}-\bar{t})^{2}}},\hat{t} = f(X;\theta_{f})
\end{equation}
Specifically, the ${\rm MCC}_{Xt}$ has distinct implications under different $f(X;\theta_{f})$: the ${\rm MCC}^{line}_{Xt}$ based on the linear regression model $f^{line}(X;\theta_{f})$ to measure the linear correlation between the covariates and the treatment variables; otherwise, the ${\rm MCC}^{nonL}_{Xt}$ based on the nonlinear model $f^{nonL}(X;\theta_{f})$ to measure the nonlinear correlation. In this paper, we chose the decision tree as $f^{nonL}(X;\theta_{f})$ since the algorithm is a non-parametric nonlinear algorithm that involves less human settings and facilitates implementation. Naturally, other nonlinear regression methods are applicable. 

Regarding inference accuracy, different from the saturation evaluation of discrete treatment, which evaluates the differential effect of each treatment with respect to other treatments, the marginal treatment effect function (MTEF) is used to measure the causal effect in the case of the continuous treatment\cite{Kreif2015Evaluation}. As demonstrated in Eq. (\ref{equ3}), the MTEF indicates the causal effect of a perturbation at a particular treatment level on the expected counterfactual outcome for all samples. In the other word, the MTEF captures the marginal change in the outcome variable caused by the treatment variables at a particular level in a differential form.
\begin{equation}
\label{equ3}
{\rm MTEF}(t) = \frac{\mathbb{E}[y_{i}(t)] - \mathbb{E}[y_{i}(t - \Delta t)]}{\Delta t}
\end{equation}
Next, we measure the accuracy of counterfactual inference by comparing the ${\rm MTEF}^{pre}$ predicted by the comparison model with the true ${\rm MTEF}^{true}$, that is the rooted mean squared error of MTEF shown in Eq.~(\ref{equ21}). 
\begin{equation}
\label{equ21}
\epsilon_{{\rm MTEF}} = \sqrt{\frac{1}{n}\sum_{i=1}^{n}({\rm MTEF}^{true}_{i}-{\rm MTEF}^{pre}_{i})^{2}}
\end{equation}
\subsection{Experimental Design}
In this paper, the proposed model is compared with various state-of-the-art counterfactual inference models for continuous treatment variables that are based on reweighting or deep networks:  including ICPW\cite{2000Marginal} , GBM\cite{2015A}, CBGPS and npCBGPS\cite{2018Covariate}, GAD\cite{2021continuous}, DRNets\cite{2019LearningC}, and SCIGAN\cite{SCIGAN}.

The datasets for the aforementioned four scenarios are divided into training/validation/test sets according to the percentages of 60/20/20. Further, as obtaining consistent causal conclusions is one of the primary objectives of causal inference, we treat sampling data greater than eighty percent of the quantile of the treatment variables as test set and the rest as training and validation set. This partitioning enables us to measure the generalization performance of the comparison model by evaluating its ability to perform well outside the training domain.

To ensure a fair comparison of the comparison models, a systematic grid search approach is adopted to select the optimal hyperparameters. This entails selecting the best hyperparameters for each model from a predefined range based on its performance on the validation sets. Subsequently, we evaluate the chosen models 100 times to record the mean and standard error of the evaluation metrics. For the above evaluation metrics, we give the values in contexts including training and test sets.

\subsection{Results and Discussion}
\subsubsection{Evaluation of the de-confounding effect}
\begin{figure*}[!t]
\centering
\subfloat[Scenarios with linear $f_{t}^{line}$.]{\includegraphics[width=0.5782\textwidth]{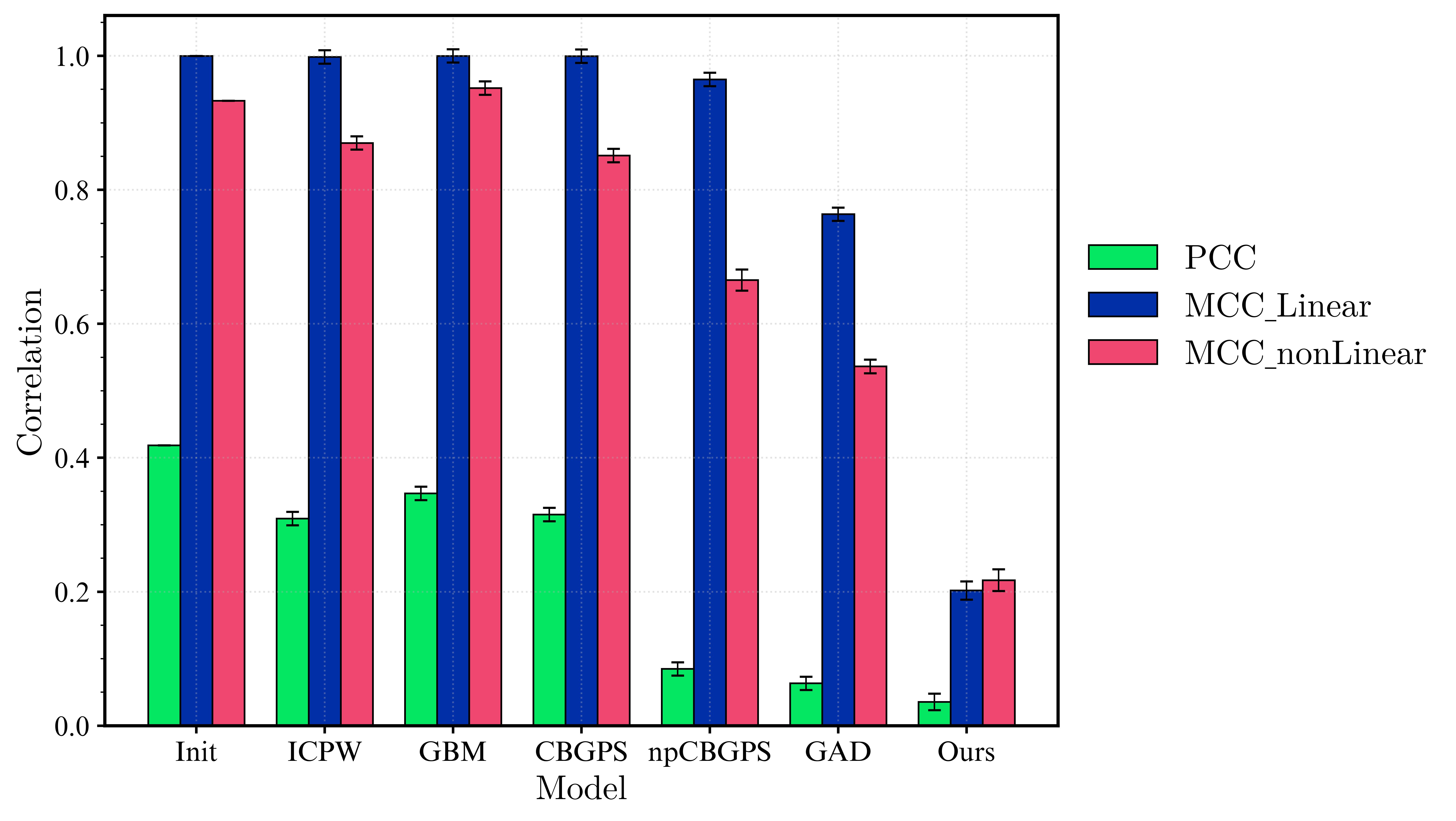}%
\label{fig:Case1}}
\hfil
\subfloat[Scenarios with nonlinear $f_{t}^{nonL}$.]{\includegraphics[width=0.4118\textwidth]{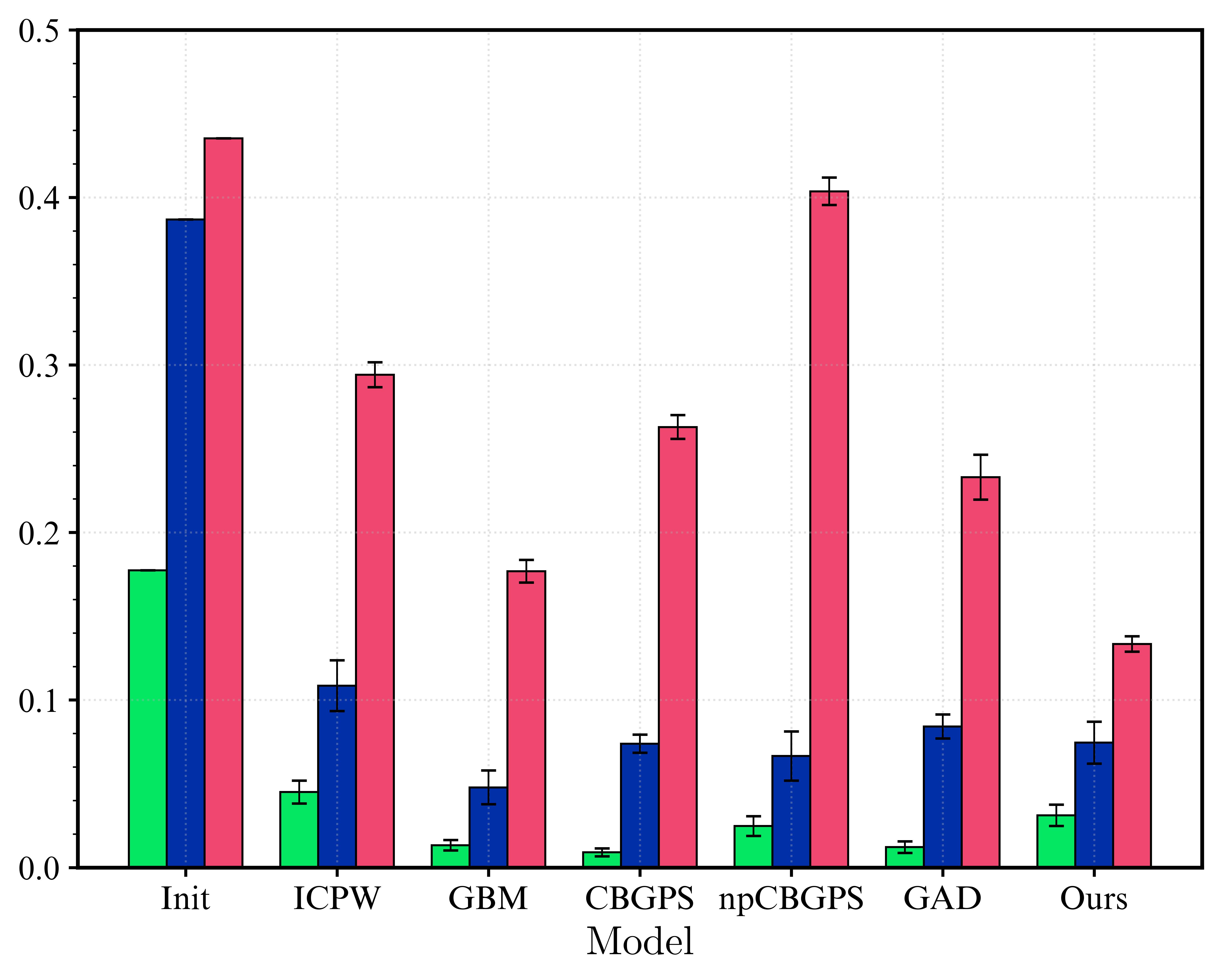}%
\label{fig:Case2}}
\caption{Comparison of the variation of correlation between treatment variables and covariates.}
\label{FIG:3}
\end{figure*}

 Fig.~\ref{FIG:3} illustrates the changes in correlations $c_{Xt}$ between the treatment variables and the covariates in the synthetic datasets processed by various comparative models. Fig.~\ref{fig:Case1} depicts scenarios with linear $f_{t}^{line}$, while Fig.~\ref{fig:Case2} illustrates scenarios with nonlinear $f_{t}^{nonL}$. Especially, the results of DRNets and SCIGAN are not included in this evaluation because these models lack de-confounding settings for continuous treatment variables. For the linear scenarios of $f_{t}^{line}$, the correlation $c_{Xt}$ in the source data is significant, as evidenced by both ${\rm MCC}^{line}_{Xt}$ and ${\rm MCC}^{nonL}_{Xt}$ being greater than 0.9. It can be observed from Fig.~\ref{fig:Case1} that various methods reduce the correlations $c_{Xt}$, with the proposed model reducing both linear and nonlinear correlations the most. Meanwhile, the DRL model significantly outperforms other state-of-the-art methods in reducing nonlinear correlation ${\rm MCC}^{nonL}_{Xt}$. For the nonlinear scenarios of $f_{t}^{nonL}$, the source data exhibits an insignificant linear correlation, indicated by both the PCC and ${\rm MCC}^{line}_{Xt}$ being less than 0.5. Consequently, the potential for decreasing the linear association between variables is restricted, resulting in comparable performance among the different methods. In contrast, the DRL model demonstrated superior performance in reducing nonlinear associations. There are two primary reasons for this phenomenon: (i) The reduction of nonlinear associations of the treatment variables with the covariates is not involved in the objectives of these methods except for GAD; (ii) In GAD, sample reweighting is essentially the process of learning a linear representation for each sample. While the nonlinear representational learning ability of DRL is more prominent than that of learning pure sample weights.

\subsubsection{Evaluation of the counterfactual inference accuracy}
Table~\ref{tab:2} presents the counterfactual inference performance of various comparison models under four pre-defined scenarios on both training and test sets. It can be realized that the performance of the various methods is consistent in the training and test sets under the linear scenarios of $f_{y}^{line}$ (Scenario A and C). The reason is that these methods make the correct assumption of linearity with respect to $f_{y}^{line}$. Therefore, for Scenario A and C, we report the results in one column. However, under the nonlinear scenarios of $f_{y}^{nonL}$ (Scenario B and D), the performance of the other comparison methods decreases significantly on test sets, and in contrast, the proposed model outperforms. Where the GAD only considers the correlation between covariates and treatment variables, ignoring the complex mapping relationships between covariates and outcome variables. Furthermore, deep network-based DRNets and SCIGAN, which are designed for dose-response models with discretized continuous doses under multiple treatment variables, still rely on achieving confounding balance under a discrete treatment and are not suitable for single continuous variable. In summary, the DRL model achieves optimal inference accuracy in both the training and test sets across the four pre-defined scenarios, which is consistent with the preceding step in which the model produced superior de-confounding representations of the covariates.
\begin{table*}[h]
  \centering
  \caption{The rooted mean squared error on MTEF $\epsilon_{{\rm MTEF}}$ of various methods.}
  \label{tab:2}
  \resizebox{\linewidth}{!}{
  \begin{tabular}{c@{}c|cc|c|cc}
    \toprule
    \multirow{3.8}{*}{Methods} &  \multicolumn{6}{c}{{\bfseries Scenarios}}\\
    \cmidrule{2-7}
    & \multicolumn{1}{c}{A($f_{t}^{line}$ and $f_{y}^{line}$)} & \multicolumn{2}{|c|}{B($f_{t}^{line}$ and $f_{y}^{nonL}$)} & \multicolumn{1}{c}{C($f_{t}^{nonL}$ and $f_{y}^{line}$)} & \multicolumn{2}{|c}{D($f_{t}^{nonL}$ and $f_{y}^{nonL}$)} \\
    \cmidrule{2-7}
    & training \& test &  training & test & training \& test &  training & test \\
    \midrule
    \multicolumn{1}{c|}{MSM}  & $0.80\pm0.01$ & $0.11\pm0.01$ & $0.30\pm0.01$ & $14.66\pm0.43$ & $0.35\pm0.11$ & $1.21\pm0.14$\\
    \multicolumn{1}{c|}{ICPW}  & $0.76\pm0.01$ & $0.08\pm0.08$ & $0.37\pm0.01$ & $3.75\pm0.54$ & $0.44\pm0.17$ & $0.48\pm0.13$\\
    \multicolumn{1}{c|}{GBM}  & $0.71\pm0.01$ & $0.08\pm0.01$ & $0.16\pm0.01$ & $1.01\pm0.47$ & $0.44\pm0.17$ & $0.45\pm0.16$\\
    \multicolumn{1}{c|}{CBGPS} & $0.80\pm0.01$ & $0.10\pm0.01$ & $0.39\pm0.02$ & $0.36\pm0.23$ & $0.33\pm0.14$ & $0.71\pm0.15$\\
    \multicolumn{1}{c|}{npCBGPS} & $0.72\pm0.01$ & $0.05\pm0.05$ & $0.20\pm0.14$ & $1.84\pm0.53$ & $0.72\pm0.19$ & $0.79\pm0.15$\\
    \multicolumn{1}{c|}{GAD} & $0.69\pm0.01$ & $0.06\pm0.01$ & $0.29\pm0.03$ &  $0.33\pm0.19$ & $0.36\pm0.09$ & $0.74\pm0.13$\\
    \multicolumn{1}{c|}{DRNets} & $1.22\pm0.01$ & $0.06\pm0.01$ & $0.32\pm0.02$ &  $2.23\pm0.34$ & $0.36\pm0.14$ & $0.56\pm0.16$\\
    \multicolumn{1}{c|}{SCIGAN} & $1.54\pm0.02$ & $0.08\pm0.01$ & $0.10\pm0.07$ &  $0.75\pm0.04$ & $0.38\pm0.05$ & $0.41\pm0.14$\\
    \multicolumn{1}{c|}{{\bfseries Ours}} & \bm{$0.49\pm0.01$} & \bm{$0.04\pm0.01$} & \bm{$0.07\pm0.01$} & \bm{$0.10\pm0.02$} & \bm{$0.19\pm0.09$} & \bm{$0.22\pm0.10$}\\
  \botrule
\end{tabular}
}
\end{table*}

\section{Case Analyses}
\subsection{Background and Data Description}
Red cell distribution width (RDW) is a metric of erythrocyte size variability and has been identified as a prognostic indicator for patient mortality\cite{2012Red,Jae2013Red,2014TheInt}. However, the precise mechanisms linking RDW to patient mortality remain inadequately understood. To address this knowledge gap, we conduct a case study that aims to investigate the detailed causal relationship between RDW and mortality using the MIMIC \uppercase\expandafter{\romannumeral3} database. This medical record database comprises an extensive, independent, and unselected population of patients admitted to the intensive care unit (ICU)\cite{MIMIC1}. 

The MIMIC \uppercase\expandafter{\romannumeral3} database includes clinical variables such as demographics (age, gender), highly granular physiologic data captured by the bedside monitors, medications administered and procedures performed, chronic disease diagnoses as represented by the International Classification of Diseases ICD-9 codes, as well as laboratory results (complete blood count, serum chemistries, and microbio-logic data)\cite{MIMIC1,MIMIC2}. Additionally, survival outcome data after hospital discharge is obtained from the Social Security database. Following the \cite{2012Red}, we include all adult patients who were admitted to ICU floors and had the RDW measurements on admission, resulting in a total of 47,525 medical records. The experiment's continuous treatment variable is RDW, with a research range of 12\% to 18\%, and the outcome variable is mortality within one year of hospital discharge. The covariates considered in the analysis include gender, admission age, duration of hospitalization, duration of ICU stay, simplified Acute Physiology Score (SAPS) \uppercase\expandafter{\romannumeral3}, hematocrit, and presence of complications. 

\subsection{Applications and Analyses}
First, we report the correlation between the representations of the covariates generated by the proposed method and the treatment variables, as shown in Table~\ref{table:3}. It is obvious from Table~\ref{table:3} that the DRL model reduces the metrics PCC, ${\rm MCC}^{line}_{Xt}$ and ${\rm MCC}^{nonL}_{Xt}$ between the original covariates and the treatment variables. Furthermore, since the mortality prediction is a category imbalance problem, we employ the area under the receiver operator characteristic curve (AUC) to gauge the model's predictive performance. And the AUC score of the DRL model on the MIMIC \uppercase\expandafter{\romannumeral3} dataset can reach 0.83, which is a satisfactory level.
\begin{table}[h]
\centering
\caption{The practical performance of the proposed model on the MIMIC \uppercase\expandafter{\romannumeral3} dataset.}
\label{table:3}
\begin{tabularx}{\textwidth}{>{\centering\arraybackslash\hsize=.28\hsize}X |>{\centering\arraybackslash\hsize=.18\hsize}X |>{\centering\arraybackslash\hsize=.18\hsize}X |>{\centering\arraybackslash\hsize=.18\hsize}X }
\toprule
 Metrics & PCC & ${\rm MCC}^{line}_{Xt}$ & ${\rm MCC}^{nonL}_{Xt}$  \\
 \midrule
 MIMIC \uppercase\expandafter{\romannumeral3} Data & 0.18 & 0.44 & 0.46  \\ 
 Generated Data  & 0.09 & 0.29 & 0.31  \\
 \botrule
\end{tabularx}
\end{table}

\begin{figure}[h]
    \centering
    \includegraphics[width=0.6\textwidth]{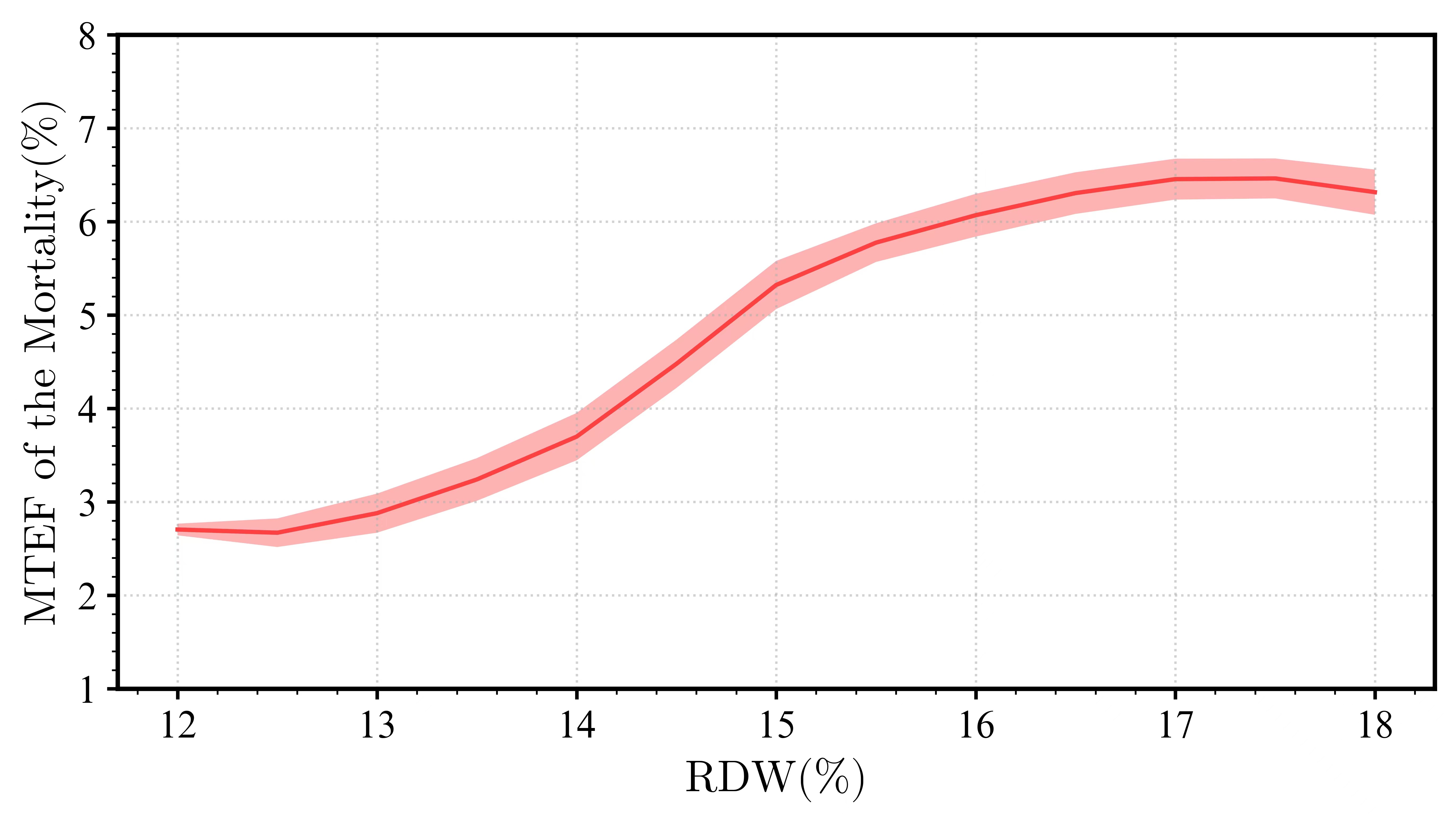}
    \caption{The average MTEF of RDW on the mortality within one year of hospital discharge.}
    \label{RDW}
    \end{figure}

With the acquired de-confounding representations and precise inference, we evaluated the MTEF regarding mortality at various RDW levels. As illustrated in Fig.~\ref{RDW}, the marginal causal effect MTEF on mortality increases progressively with increasing RDW. This phenomenon suggests that RDW is a risk factor for mortality in the range of 12\% to 18\%, and the marginal increase in mortality for a given increment in RDW is more significant at higher RDW levels. This investigation has practical implications for clinical practice, as it enables more precise risk stratification and timely treatment for patients.
\section{Conclusion}
The dependence between confounding covariates and treatment variables presents a primary challenge to existing methods based on sample reweighting. These methods often prioritize the elimination of linear dependencies while neglecting nonlinear correlations. On the other hand, some methods exhibit high training complexity, which hinders their practical application. In this paper, we propose a counterfactual inference framework DRL for continuous treatment by simultaneously reducing the linear and nonlinear dependence between the de-confounding representations of covariates with treatment variables. We integrate an adversarial network for de-confounding representations and a counterfactual inference network to serve the dual purposes of de-confounding and counterfactual inference. The DRL eliminates as much spurious correlations result from confounders as possible, and outperforms state-of-the-art counterfactual inference models for continuous treatment variables on the synthetic datasets. Furthermore, we present a real-world application of the proposed framework to predict mortality rates based on the
medical database MIMIC \uppercase\expandafter{\romannumeral3}. Our findings reveal a detailed causal association between RDW and mortality, highlighting the practical utility of our model.

\bibliography{mybib}

\end{document}